# RALTPER: A Risk-Aware Local Trajectory Planner for Complex Environment with Gaussian Uncertainty

Cheng Chi

*Abstract*—In this paper, we propose a novel Risk-Aware Local Trajectory Planner (RALTPER) for autonomous vehicles in complex environments characterized by Gaussian uncertainty. The proposed method integrates risk awareness and trajectory planning by leveraging probabilistic models to evaluate the likelihood of collisions with dynamic and static obstacles. The RALTPER focuses on collision avoidance constraints for both the ego vehicle region and the Gaussian-obstacle risk region. Additionally, this work enhances the generalization of both vehicle and obstacle models, making the planner adaptable to a wider range of scenarios. Our approach formulates the planning problem as a nonlinear optimization, solved using the IPOPT solver within the CasADi environment. The planner is evaluated through simulations of various challenging scenarios, including complex, static, mixed environment and narrow single-lane avoidance of pedestrians. Results demonstrate that RALTPER achieves safer and more efficient trajectory planning particularly in navigating narrow areas where a more accurate vehicle profile representation is critical for avoiding collisions.

## I. INTRODUCTION

With the rapid advance of autonomous driving technology, ensuring vehicles can plan and execute trajectories safely and efficiently in complex and dynamic environments becomes increasingly important. Current trajectory planning methods mainly focus on the safety and feasibility of paths, but considering environmental uncertainty generating a risk-aware local trajectory remains a challenging problem.

Traditional sampling-based or search-based algorithms[1][10] have been widely applied in path planning. These algorithms can generate optimal paths in known environments but show their limitations in dynamic and uncertain environments. The Hybrid A* algorithm[8] is often used in the initial search of two-stage planning due to its ability to satisfy vehicle kinematic constraints.

To address environmental uncertainty, some studies have introduced probabilistic models, such as Gaussian distributions and Monte Carlo method, to predict the positions and trajectories of obstacles[11][16]. These methods enhance the safety of path planning by establishing uncertainty distributions for obstacle positions. In [15], Monte Carlo method evaluates the collision probability of a given trajectory, but the large number of samples required may not meet the real-time demands of task. In [16], on account of Gaussian motion and sensing uncertainty, a truncated estimated priori state distributions method is proposed to estimate path collision probability based on the robot's prior probability distribution. In [15], a linear variation function of Gaussian uncertainty for obstacles is designed, covering the vehicle model with multiple disks and expanding the risk area. However, in narrow environments, overly conservative model construction may lead to planning failures

As the pioneering study, [17] proposed the method of risk-aware planning. The concept of risk contours was proposed in [18] representing the risk contour problem as a risk constraint problem and using the theory of moments and nonnegative polynomials to provide a convex optimization in the form of sum of squares optimization. In [19] considering non-Gaussian uncertainties, motion planning for stochastic nonlinear systems is completed by using offline-constructed discrete-time motion primitives and their corresponding continuous-time tubes. Probabilistic surrogate reliability and risk contours were introduced in [20] with the minimal covering disk calculation for the vehicle model. Subsequent research has produced more algorithms based on risk-bounded verification, but they focus on calculating risk contours for non-convex obstacles and find it challenging to use sums of squares for real-time online planning.

Extensive research [21][22]on MPC has promoted the development of trajectory planning that meets system kinematic constraints. Adding reasonable safety constraints for obstacles is crucial in these studies. [23] describes both the vehicle and obstacle models using hyperplanes and constructs a minimal distance function, transforming it into convex constraints of the MPC problem using the Lagrangian dual principle. Subsequent studies [24] and [25] added warm-start elements and applied the methods in the Apollo system. However, all of these methods assume deterministic parameters and lack measures for obstacle uncertainty. Therefore, it's highly valuable and significant to design a risk-aware MPC local trajectory planner.

This paper proposes a new nonlinear MPC method for probabilistic safe local trajectory planning of autonomous vehicles in uncertain environments. The research aims to develop a local trajectory planner capable of generating risk-aware, aggressive yet safe trajectories under vehicle kinematic constraints. First, the planner considers obstacle uncertainties

Cheng Chi, with the School of Instrument Science and Engineering, Southeast University, Nanjing 210096, China (e-mail: 220223321@seu.edu.cn).

using the Kalman filter method, providing a reasonable expression of uncertainties. Based on this, Gaussian distribution models was applied to identify risk areas. Finally, using an existing global path planning method (Hybrid A* in this paper) as the initial feasible solution of the optimal control problem, safety distance constraints between the Gaussian risk distribution area and the hyperplane-represented vehicle model were added to perform nonlinear planning and determine the optimal collision-free trajectory. The planner's workflow is shown in Fig. 1. Compared with existing literature, the main contributions of this work are summarized as follows:

- Developing a risk-aware local trajectory planner suitable for autonomous vehicles in dynamic and uncertain environments. Compared with the traditional local planners, risk awareness is added to make trajectory planning safer and more reliable;
- A reasonable derivation of Gaussian uncertainty risk areas and a reasonable propagation method for uncertainty risks are provided by calculating the covariance matrix of the Kalman filter;
- The minimum distance function between the convex polygon formed by hyperplanes and circular (elliptical) shapes is calculated, enhancing the generalization of the vehicle and obstacle models, thereby improving adaptability to different scenarios;
- By applying the Lagrangian dual conditions, the minimum distance function is transformed into a constraint in the MPC problem. Additionally, the Gaussian uncertainty is converted into deterministic constraints within the optimization problem, ensuring that the risk associated with trajectory planning remains below the given acceptable threshold.

## II. PROBLEM STATEMENT

The goal of the risk-aware local trajectory planner is to generate risk-aware trajectories within environments containing obstacles with Gaussian uncertainty. These trajectories should be more aggressive while remaining within acceptable risk levels, and must comply with the vehicle's own kinematic constraints. This approach can be broadly applied to various scenarios requiring local planning, such as trajectory planning in complex and mixed-obstacle environments in open spaces, or local planning in narrow lanes during autonomous parking.

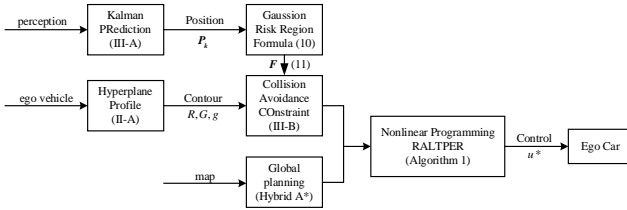

**Fig. 1.** This is a sample of a figure caption.

### A. Ego Vehicle Profile Representation

In the field of robotics, particularly in the navigation and control of autonomous vehicles, accurately representing the vehicle's profile is crucial for tasks like collision avoidance, path planning, and environmental interaction. Using hyperplane constraints is an effective method to represent the vehicle's profile. By defining a set of hyperplanes, a geometric shape can be described, such as a rectangle using four hyperplanes to approximate the physical dimensions of the vehicle. This can be expressed through inequalities in the form $Gx \leq g$, where $G$ is a coefficient matrix, and $g$ is a constant vector. To achieve a more precise description, the number of hyperplanes can be increased to more finely capture the vehicle profile.

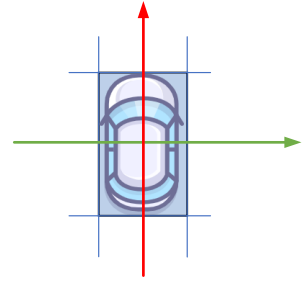

**Fig. 2.** Defines a ego vehicle profile region centered at the origin

A normal ego vehicle region are shown in Fig. 2, where the vehicle's length is $L$ and width is $W$. The coefficient matrix $G = [1,0,-1,0;0,1,0,-1]^T$ and constant vector $g = [L/2, W/2, L/2, W/2]^T$ are defined accordingly.

### B. Local Trajectory Planning

Local planning refers to the process of generating feasible paths or trajectories for the vehicle within a dynamically changing environment. The primary goal is to ensure safe and efficient movement from the vehicle's current position to the target position, while avoiding obstacles and adhering to the vehicle's kinematic constraints.

The local planning problem can be formulated as an optimization problem, comprising the following components:

**Objective Function**: The objective is to minimize a cost function, which is a weighted sum of metrics in the trajectory planning task, typically including path length, cost time, energy, and safety distances, among others.

$$J = \sum_{i=1}^{n_J} J_i \quad (1)$$

**Constraints**: The optimization problem is subject to various constraints, including kinematic constraints, safety constraints, and states space constraints.

1) Kinematic Constraints

The vehicle's kinematic principles can be defined as

$$x(k+1) = f(x(k), u(k)) \quad (2)$$

These kinematic constraints must be satisfied throughout the planning time horizon $[t_0, t_N]$. The state variables $x = [x \ y \ \theta]^T$ include the vehicle's position in the $x$ and $y$ directions and its current orientation. The control variables $u = [v \ \omega]^T$ include the vehicle's speed and angular velocity.

2) Safety Constraints

The vehicle model is represented using hyperplanes, while other traffic participants, such as pedestrians, can be covered by disks. The vehicle's region is represented as $\mathbb{E}$, and the obstacle regions as $\chi$. We require that the safety constraints between the vehicle and the obstacles as follow.

$$\mathbb{E} \cap \chi_i(k) = \varnothing \quad (3)$$

Where $i = \{1, 2, \ldots, n_o\}$, $n_o$ represents the number of obstacles. This implies that the vehicle's region should not intersect with any obstacle regions at any time during the planning horizon.

3) State Space Constraints

Subject to the mechanical properties of the vehicle and the influence of the state space search area, the constraints must be satisfied as follow

$$\begin{aligned} x \in [\underline{x}, \overline{x}], \ \forall t \in [t_0, t_N] \\ u \in [\underline{u}, \overline{u}], \ \forall t \in [t_0, t_N] \end{aligned} \quad (4)$$

Then the trajectory planning task can be defined as the following optimal control problem

$$p^* = \min \ (1) \ \text{s.t. Conditions (2)-(4)} \quad (5)$$

Common control methods are deterministic, based on current perception, such as pure pursuit control[26]. However, these require accurate perception and prediction of obstacles. In the physical world, uncertainty is prevalent, and Gaussian uncertainty for obstacles is a common assumption. In this study, a filter-based prediction module can model the Gaussian uncertainty risk region, and a safety constraint between the vehicle model and the Gaussian uncertainty risk region is designed, constructing a risk-aware local trajectory planner.

III. RISK-AWARE LOCAL TRAJECTORY PLANNER

Inspired by [18], [20], [23], [24], [25], the expression of Gaussian risk uncertainty was transformed into an acceptable collision risk region, as explained in Section III-A. In this context, while also considering the design of collision avoidance constraints, we expanded the form of collision constraints to assess the safety constraints between the ego vehicle region and the Gaussian risk region, as detailed in Section III-B. Section III-C will introduce the overall design of the planner.

*A. Obstacle Risk Region*

Taking pedestrians as an example of obstacles, assume that the obstacle area is covered by a disk. The state of a pedestrian is defined as $s = (x, y, \theta, v, \omega)$, where $(x, y)$ represents the center position of the obstacle disk, $\theta$ indicates the current moving direction of the obstacle; $v$ represents the speed along the current moving direction, and $\omega$ is the angular velocity of the obstacle.

The motion model of the obstacle can be constructed as a Constant Velocity model, Constant Turn Rate model, or Constant Turn Rate and Velocity model. Single model or Interacting Multiple Model tracking and prediction can be performed using Extended Kalman Filter or Unscented Kalman Filter.

In each prediction step, the prediction covariance matrix $P_{k+1/k}$ is obtained at each moment, from which the variances and covariance in the x and y dimensions are extracted to form a new two-dimensional covariance matrix $P_k$. The prediction result $\hat{s}_{k+1/k}$ at this moment is the mean of the position distribution of the obstacle under Gaussian uncertainty, and the two-dimensional covariance matrix $P_k$ represents the probabilistic distribution probability.

Assuming a Gaussian distribution $\mathcal{N}(\mu, P_k)$ with a mean $\mu$ and covariance matrix $P_k$, and given an acceptable collision probability $\alpha$, the confidence interval for the obstacle position is $100(1-\alpha)\%$. On the assumption that a simple case where $\mu = [0; 0]$, $P_k = diag([\sigma_x^2, \sigma_y^2])$ and theoretical derivation is provided as follows.

If the above assumptions hold, the region can be represented as

$$(\frac{x}{\sigma_x})^2 + (\frac{x}{\sigma_y})^2 = c \quad (6)$$

The parameter $c$ is determined by directly integrating the density function.

$$\oint_D \frac{1}{2\pi \sigma_x \sigma_y} e^{-\frac{1}{2}\left[(\frac{x}{\sigma_x})^2 + (\frac{x}{\sigma_y})^2\right]} dxdy = 1 - \alpha \quad (7)$$

Where $x = r\sigma_x \cos\theta$ and $y = r\sigma_y \sin\theta$, then

$$\left|\frac{d(x,y)}{d(r,\theta)}\right| = \left|\begin{matrix} \sigma_x \cos\theta & -r\sigma_x \sin\theta \\ \sigma_y \sin\theta & r\sigma_y \cos\theta \end{matrix}\right| = r\sigma_x \sigma_y \quad (8)$$

We obtain the expression about $\alpha$

$$c = -2\ln\alpha \quad (9)$$

Therefore, the region with a probability distribution less than $100(1-\alpha)\%$ is provided as follow

$$\chi = \left\{ (x, y) \mid (\frac{x}{\sigma_x})^2 + (\frac{x}{\sigma_y})^2 \leq -2\ln\alpha \right\} \quad (10)$$

Similarly, it can be proven that for non centralized binary normal distributions, the above derivation can be done after centralization.

**Illustrative Example 1**: Consider the following illustrative example. We assume that the position distribution of a obstacle at a moment $t_k$ has Gaussian uncertainty. Given that the position distribution of the obstacle in terms of its *x* and *y* coordinates follows Gaussian distribution with mean $\mu$ and covariance matrix $P_{t_k}$, we can describe this mathematically as follows:

The position of the obstacle can be represented by a two-dimensional vector. The mean position of the obstacle is given by the two-dimensional vector $\mu = [0; 0]$. The covariance matrix $P_{t_k}$ is a 2×2 matrix that describes the covariance

between the *x* and *y* coordinates of the obstacle's position. It is given by:

$$\boldsymbol{P}_{t_k} = \begin{bmatrix} 2 & 0.5 \\ 0.5 & 1 \end{bmatrix}$$

Fig. 3 illustrates the risk region under different scenarios where the collision probability $\alpha$ is less than 1%, less than 5%, and less than 10%, considering the Gaussian uncertainty in the position of the obstacle. Indeed, as the collision probability threshold decreases, the risk region tend to become more conservative.

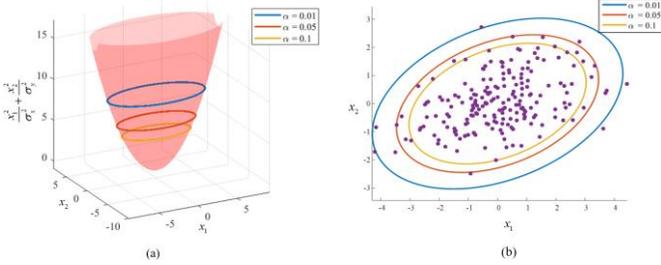

**Fig. 3.** Risk region under different collision probability $\alpha$

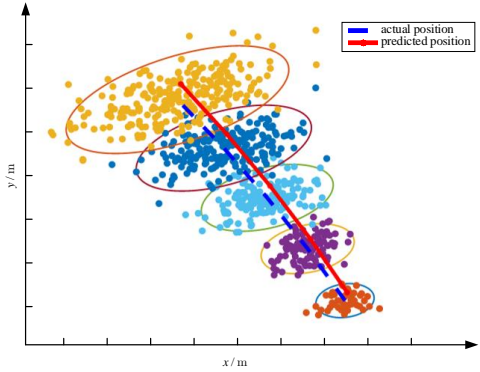

**Fig. 4.** Dynamic obstacle prediction risk region with collision risk less than 10%

**Illustrative Example 2**: Consider the following illustrative example. I assume there exists a moving point in space, and the motion of this point conforms to a constant velocity model. The state transition equation of the CV model is expressed as

$$\boldsymbol{S}_k = \boldsymbol{S}_{k-1} + \begin{bmatrix} v\cos(\theta)T \\ v\sin(\theta)T \\ \omega T \\ 0 \\ 0 \end{bmatrix}$$

The motion parameters of the point are set as follows: velocity $v$=1.8 m/s, angular velocity $\omega$=1.5 rad/s, with a sampling time interval of 0.1 s. The standard deviation of noise in the state update process is 0.05m, and the standard deviation of noise in the measurement process is 0.1m. Fig. 4 shows that we select a moment $t_k$ during the process, performs a five-steps state prediction for the point, and generates a risk region where the collision probability for the point is less than 10%.

*B. Collision Avoidance Constraint*

Based on the description in section III-A, the Gaussian uncertainty risk region of the obstacle is defined as $\chi_k^i$, representing the uncertainty risk region of the *i*th obstacle at time *k*. A point within the obstacle risk region is defined as $o = [x, y]^T$, which should satisfy.

$$\boldsymbol{o}^T \boldsymbol{F} \boldsymbol{o} \le 1 \quad (11)$$

Where $\boldsymbol{F} = diag([1/(-2\ln\alpha \cdot \sigma_x^2), 1/(-2\ln\alpha \cdot \sigma_y^2)])$. Obviously, it is a positive-definite matrix.

The region of the ego vehicle at any time can be obtained by rotating and translating a pre-defined region centered at the origin

$$\mathbb{E} = \boldsymbol{R}(x)\mathbb{B} + \boldsymbol{t}(x), \quad \mathbb{B} := \{\boldsymbol{y} : \boldsymbol{G}\boldsymbol{y} \le \boldsymbol{g}\} \quad (12)$$

$$\chi = \boldsymbol{A}(x)\mathbb{O} + \boldsymbol{b}(x), \quad \mathbb{O} := \{\boldsymbol{o} : \boldsymbol{o}^T \boldsymbol{F} \boldsymbol{o} \le 1\} \quad (13)$$

where $\mathbb{B} \subset \mathbb{R}^2$ is the pre-defined rectangular region centered at the origin, $\boldsymbol{R} \in \mathbb{R}^{2\times 2}$ is the rotation matrix, and $\boldsymbol{t} \in \mathbb{R}^2$ is the translation vector. Similarly, where $\mathbb{O} \subset \mathbb{R}^2$ is the pre-defined elliptical region centered at the origin, $\boldsymbol{A} \in \mathbb{R}^{2\times 2}$ is the rotation matrix, and $\boldsymbol{b} \in \mathbb{R}^2$ is the translation vector.

During the process of computing the current state of the ego vehicle $x_0$ to the target state $x_G$, the goal is to maintain a minimum distance from the obstacle. The minimum distance between the ego vehicle region and the obstacle risk region can be defined as

$$dist(\mathbb{E}, \chi) = \min_{e,o} \|e - o\|$$
$$s.t. \quad e \in \mathbb{E}, \ o \in \chi \quad (14)$$

Fig. 5 illustrates the minimum distance between the ego vehicle region and the Gaussian risk region. Based on the above transformation, the minimum distance function can be rewritten as:

$$dist(\mathbb{E}, \chi) = \min_{e', o'} \|\boldsymbol{R}e' + \boldsymbol{t} - (\boldsymbol{A}o' + \boldsymbol{b})\|$$
$$s.t. \quad \boldsymbol{G}e' \le \boldsymbol{g} \quad (15)$$
$$\boldsymbol{o}'^T \boldsymbol{F} \boldsymbol{o}' \le 1$$

This equation involves two optimization variables. Introducing a new optimization variable $\omega$, the problem can be rewritten as

$$dist(\mathbb{E}, \chi) = \min_{\omega} \|\omega\|$$
$$s.t. \quad \boldsymbol{R}e' + \boldsymbol{t} - (\boldsymbol{A}o' + \boldsymbol{b}) = \omega$$
$$\boldsymbol{G}e' \le \boldsymbol{g} \quad (16)$$
$$\boldsymbol{o}'^T \boldsymbol{F} \boldsymbol{o}' \le 1$$

This is a standard optimization problem. By introducing the Lagrange function, the constrained optimization is transformed into an unconstrained optimization:

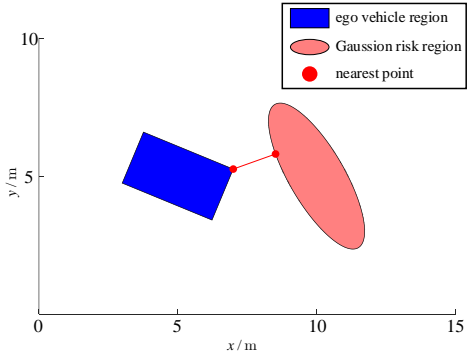

**Fig. 5.** The minimum distance between the ego vehicle region and the Gaussian risk region

$$L(\omega,\lambda,\mu,\gamma) = \|\omega\| + \lambda^T \left[Re' + t - (Ao' + b) - \omega\right] \\ + \mu^T (Ge' - g) + \gamma (o'^T Fo' - 1) \quad (17)$$

The Lagrange multiplier $\lambda, \mu, \gamma$ should satisfy $\mu \geq 0, \gamma \geq 0$. The Lagrange dual function is provided as follow

$$g(\lambda,\mu,\gamma) = \inf_\omega \left[\|\omega\| - \lambda^T \omega\right] + \inf_{e'} \left[\lambda^T Re' + \mu^T Ge'\right] \\ + \inf_{o'} \left[-\lambda^T Ao' + \gamma o'^T Fo'\right] \\ + \lambda^T (t - b) - \mu^T g - \gamma \quad (18)$$

Given that $\inf_\omega \left[\|\omega\| - \lambda^T \omega\right]$ is defined by a norm, and since the norm is non-negative, $\lambda^T \omega$ is an inner product. When $\|\lambda\| \leq 1$, $\|\lambda\|$ is constrained within a unit ball. According to the Cauchy-Schwarz inequality, for any $\omega$ and $\lambda$, we have $|\lambda^T \omega| \leq \|\lambda\| \cdot \|\omega\|$. Thus, when $\|\lambda\| \leq 1$, we get $-\|\lambda\| \cdot \|\omega\| \leq \lambda^T \omega \leq \|\lambda\| \cdot \|\omega\|$. Hence, $\inf_\omega \left[\|\omega\| - \lambda^T \omega\right] \geq 0$. Therefore, a lower bound exists, which is 0. When $\|\lambda\| \geq 1$, it cannot be guaranteed that $\|\omega\| - \lambda^T \omega$ is non-negative, and thus no lower bound exists.

The term $\inf_{e'}\left[\lambda^T Re' + \mu^T Ge'\right]$ can be rewritten as $\inf_{e'}\left[(\lambda^T R + \mu^T G)e'\right]$, which is a linear function concerning $e'$. If it can be proven that this expression has a lower bound, let the coefficients of $e'$ in the terms $\lambda^T R + \mu^T G = 0$ be equal to zero, then the expression has a lower bound, which is 0.

The term $\inf_{o'}\left[-\lambda^T Ao' + \gamma o'^T Fo'\right]$ is a quadratic term. The lower bound of this expression is found by solving the first-order derivative equal to zero

$$-\lambda^T A + 2\gamma Fo' = 0 \quad (19)$$

That is, when $o' = (\lambda^T A / 2\gamma F)^T$, the lower bound of $\inf_{o'}\left[-\lambda^T Ao' + \gamma o'^T Fo'\right]$ is $-\|\lambda^T A\|^2 / 4\gamma F^T$.

Lagranian duality issues can be rephrased as follows

$$d^* = \max_{\lambda,\mu,\gamma} \; -\|\lambda^T A\|^2 / 4\gamma F^T + \lambda^T (t - b) - \mu^T g - \gamma \\ s.t. \quad \|\lambda\| \leq 1 \\ \lambda^T R + \mu^T G = 0 \\ \mu \geq 0, \gamma \geq 0 \quad (20)$$

The solution of the original problem is defined as $p^*$, and the solution of the Lagrange dual problem is defined as $d^*$. Based on the duality relationship, we know that the solution to the original problem is always greater than or equal to the solution of the dual problem, $p^* \geq d^*$. We set a certain value for the solution of the dual problem and obtain the relationship $p^* \geq d^* \geq d_{\min}$. If we want to ensure that the minimum distance between the ego vehicle region and the obstacle risk region is greater than a certain value, we can add the constraint:

$$-\|\lambda^T A\|^2 / 4\gamma F^T + \lambda^T (t - b) - \mu^T g - \gamma \geq d_{\min} \\ \|\lambda\| \leq 1 \\ \lambda^T R + \mu^T G = 0 \quad (21)$$

*C. Risk-Aware Local Trajectory Planner*

First, to address the issue of large-scale jitter that occurs during the movement of actual ground robots, a new penalty term is introduced in the objective function. In general simulation environments, the motion of the vehicle is under ideal conditions, and the trajectory planning results between each control cycle tend to be consistent, without large-scale jitter. However, during the actual driving process of the robot, due to the uncertainty in the position of obstacles, there can be significant jumps in the results between two adjacent control cycles. Therefore, a jump penalty term $J_h$ is introduced in the objective function, taking the trajectory planning result of the previous control cycle as a parameter for the next control cycle, preventing significant jumps in the results. Since trajectory planning is gradual, no restrictions are added to the last step of the control step.

$$J_h = \sum_{k=1}^{N-1} \|x(k) - P(k)\| \quad (22)$$

Secondly, because the number of constraints is proportional to the number of obstacles and control steps, it can slow down the optimization problem-solving process, and even lead to unsolvable situations. Moreover, the uncertainty prediction of obstacles decreases the confidence level as the prediction time increases. Therefore, a distance penalty term from the goal is introduced in the objective function, assuming that the number of control cycle steps in trajectory planning is fixed, so that the position of each control cycle tends to the target point.

$$J_d = \sum_{k=1}^{N} \|x(k) - s_G\| \quad (23)$$

Finally, to ensure control smoothness and reduce the robot's energy consumption, a control smoothness penalty term is introduced to minimize control energy and minimize the energy difference between two control quantities.

$$J_u = \sum_{k=1}^{N} \|u(k)\| + \sum_{k=1}^{N-1} \|u(k+1) - u(k)\| \quad (24)$$

Therefore, the objective function of the risk-aware local trajectory planner is as follows

$$J(x,u) = \alpha_1 \cdot J_h + \alpha_2 \cdot J_d + \alpha_3 \cdot J_u \quad (25)$$

where $\alpha_1, \alpha_2, \alpha_3$ is the hyperparameter of the objective function.

Compared to the typical problems proposed in previous articles: (1) The hyperparameter $d_{min}$ needs to be manually set and is difficult to tune; (2) The initial and target states as constraints slow down the solving process, and in extreme cases, it is impossible to solve a feasible solution.

The advantages of this algorithm are: (1) By introducing the risk region with Gaussian uncertainty, there is no need to re-tune the hyperparameter $d_{min}$, and the safety constraint can be satisfied as long as the condition $d_{min} \geq 0$ is met within the collision risk; (2) Without hard constraints on the initial and target states, the short control cycle of the local trajectory planner greatly improves the success rate of solving. Therefore, the complete expression of the risk-aware local trajectory planner is as follows

$$\min_{x,u,\lambda,\mu,\gamma} J_N(x,u)$$

$$\text{s.t. } x(k+1) - f(x(k),u(k)) = 0$$

$$-\left\|\lambda_k^{iT} A_k^i\right\|^2 / 4\gamma_k^i F_k^{iT} + \lambda_k^{iT}(t_k^i - b_k^i) - \mu_k^{iT} g - \gamma_k^i \geq d_{min}$$

$$\left\|\lambda_k^i\right\| \leq 1$$

$$\lambda_k^{iT} R + \mu_k^{iT} G = 0$$

$$\mu_k^i \geq 0, \gamma_k^i \geq 0$$

$$x \in \mathbb{S}, u \in \mathbb{C}$$

(26)

State variables and control variables should satisfy the constraints of the state space and control space, respectively, where *i* and *k* represent the parameters of the *i*th obstacle in the *k*th control cycle.

The pseudocode for RALTPER is given in Algorithm 1. The algorithm illustrates the working process of the Risk-Aware Local Trajectory Planner (RALTPER). The algorithm begins by initializing the CasADi environment and variables, obtaining the vehicle's current position, and setting a local target. In the initial path search phase, the algorithm uses the Hybrid A* algorithm to generate an initial path based on the vehicle's profile, which serves as the foundation for subsequent optimization. Next, the algorithm performs multi-step state prediction using the Kalman filter, predicting the positions of obstacles and generating obstacle expressions and rotation matrices based on these predictions. During each iteration, the algorithm selects the path according to the current iteration count: if it is the first iteration, the initially searched path is used as the feasible solution, while in subsequent iterations, the path from the previous cycle is utilized. Then, the algorithm further optimizes the trajectory through nonlinear programming, ensuring that the vehicle's dynamic characteristics and safety constraints are taken into account. Finally, the optimized control quantity and states are output to guide the vehicle along the computed trajectory. This process is repeated in each control cycle until the vehicle successfully reaches the set local target, ensuring that trajectory planning is not only safe but also adaptable to dynamic environmental changes.

---

**Algorithm 1:** Risk-Aware Local Trajectory Planner (RALTPER)

---
**Input:** the ego vehicle profile expression $G, g$, ego vehicle rotation matrixes $R$, previous cycle $x^*$;

**Output:** The current optimal control quantity $u^*$ and states $x^*$;

1 Initialization: CasADi environment, $d$, $iter = 1$;
2 Get current location: $x0$, get local target $xs$;
3 **While** norm( $x0$, $xs$ ) < $d$ **do**
4    Initial path search (Hybrid A*), get *path*
5    Extern Kalman prediction, get obstacle expression $F$, obstacle rotation matrixes $A$;
6    **If** *iter* == 1
7      $P(k) = path$;
8    **else**
9      $P(k) = x^*$
10    **End**
11    Nonlinear programming;
12    Output the current optimal control quantity $u^*$;
13 **End**

---

## IV. EXPERIMENTAL RESULTS AND DISCUSSION

In this section, simulations are conducted to demonstrate the performance of the risk-aware local trajectory planner. Simulations were performed in two typical scenarios:

(1) trajectory planning in a complex, static, mixed environment to validate the effectiveness of safety constraints under various obstacle representations;

(2) local trajectory planning in the presence of dynamic pedestrians with Gaussian uncertainty in Narrow Single-Lane.

The nonlinear programming problem was solved using the IPOPT[27] solver in a MATLAB + CasADi[28] environment. The computations were run on Desktop PC with Intel(R) Core(TM) i7-8750H CPU, 2.20 GHz and 16.0 GB RAM, running on Windows 10.

The kinematic model of the vehicle was modeled as a general differential drive model in the experiment. However, the proposed algorithm is equally applicable to other unmanned vehicle platforms with different motion types, such as Ackermann steering. The basic vehicle parameters are: a vehicle length of 3 meters and a width of 2 meters. The relevant motion parameters are $v \in [-2, 2]$ m/s and $\omega \in [-\pi/6, \pi/6]$ rad/s.

### A. Complex, Static, Mixed Environment

In autonomous navigation tasks, developing robust path planning algorithms in complex environments with multiple obstacles is challenging. These algorithms must traverse narrow spaces, avoid obstacles of various geometric shapes,

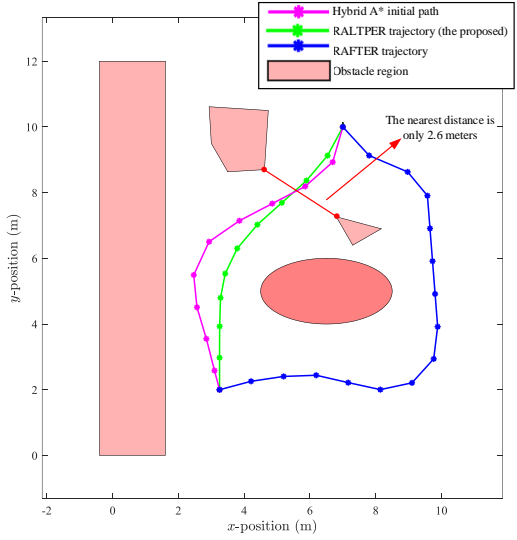 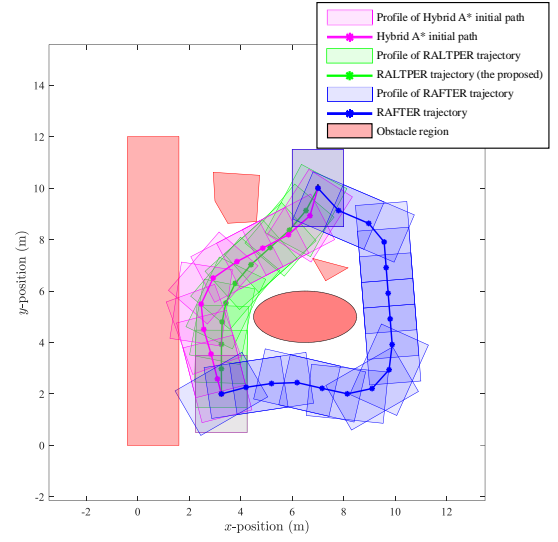

**Fig. 6.** Footprints of trajectory results generated by three algorithms

**Fig. 7.** Comparison of planned trajectories among three different algorithms

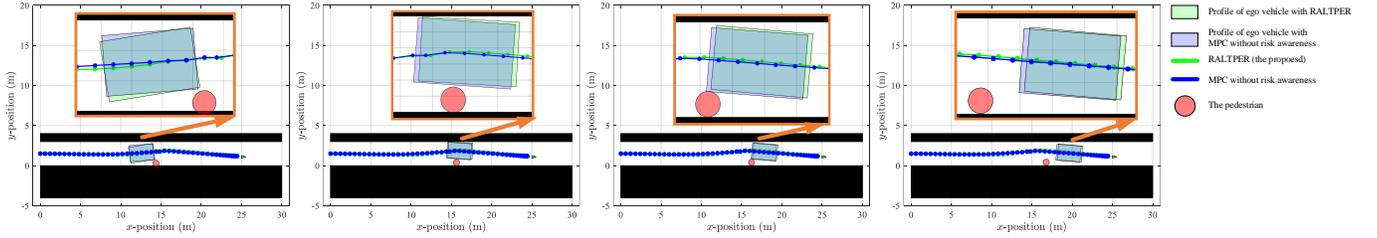

**Fig. 8.** Comparison of planned trajectories among two different algorithms in single-lane

and optimize the path to minimize travel time, energy consumption, and potential collision risks.

In a rectangular area, $x, y \in [0,12]$ m, three convex polygonal obstacles were placed, a rectangle, a triangle, and a pentagon, as well as an elliptical obstacle. The minimum distance between the obstacles is approximately 2.6 meters. The experiment compares the proposed algorithm with the RAFTER algorithm proposed by [20].

In the fig. 6, the magenta trajectory is the initial path generated by the Hybrid A* algorithm, the green trajectory is the optimized path by the proposed algorithm, and the blue trajectory is the result of the RAFTER algorithm. The initial trajectory generated by the Hybrid A* algorithm does not meet safety requirements and would collide with several obstacles. The proposed algorithm successfully avoids obstacles and performs trajectory planning in narrow spaces. According to the RAFTER method, the disk diameter is 2.6 meters, and it cannot pass through the narrowest area, thus planning a detour.

*B. Narrow Single-Lane Pedestrian Avoidance*

In urban environments, narrow one-way lanes are very common, especially in residential areas, rural roads, or passages in old parking lots. These roads typically have limited space, making it difficult for vehicles to maneuver around obstacles, and pedestrians are often present. In such scenarios, autonomous vehicles may face the following challenges:

(1) Space constraints: narrow roads with very limited space for vehicle avoidance;

(2) Pedestrians with dynamic and uncertain behavior, requiring the vehicle to be risk-aware and adjust its path in real-time.

A powerful risk-aware local trajectory planner is essential in such situations. This experiment evaluates the proposed algorithm's ability to handle uncertainty and risk in narrow spaces.

The scenario considered is a narrow one-way lane, analogous to an old parking lot environment, with a moving pedestrian in the lane. The lane is 3 meters wide, and the vehicle is 2 meters wide and 3 meters long. The pedestrian is represented by a disk with a radius of 0.3 meters. The simulation is set up such that the vehicle is driving in the center of the lane, needs to overtake the pedestrian, and reach the goal point. The pedestrian moves forward at a speed of 1 m/s, and the motion model is updated with zero-mean Gaussian noise $\mathcal{N}(0, 0.05^2)$. The vehicle's observation of the pedestrian is also subject to zero-mean Gaussian noise $\mathcal{N}(0, 0.1^2)$. The noise parameters are set to be more aggressive than in actual conditions. Comparison between the algorithm proposed in this article and MPC without risk awareness are shown in fig. 8.

The simulation results illustrate the superior performance of the proposed Risk-Aware Local Trajectory Planner (RALTPER) in navigating narrow one-way lanes while safely avoiding a moving pedestrian. The vehicle, initially positioned in the center of the lane, adeptly adjusts its trajectory in response to the pedestrian's dynamic and uncertain movements. A key feature of RALTPER is its ability to maintain a safe distance from the pedestrian throughout the entire interaction, as evidenced by the consistent gap between the vehicle and the pedestrian depicted in the simulation. The distance between the pedestrian and the vehicle's center during the vehicle's approach and departure between 3.8 seconds and 6.2 seconds is shown in Figure 9. Notably, the planner does not rush to return the vehicle to the center of the lane immediately after overtaking the pedestrian. Instead, it continues forward, ensuring the pedestrian's safety before realigning with the lane center. This demonstrates the planner's careful consideration of dynamic obstacle avoidance and its commitment to safety in real-world scenarios.

Overall, the experiment illustrates that RALTPER provides a reliable solution for autonomous navigation in constrained and dynamic environments. The planner's ability to account for both the vehicle's contour and the uncertain behavior of obstacles enhances its generalization to various real-world scenarios, making it particularly useful for urban and residential settings where narrow lanes and pedestrian interaction are common.

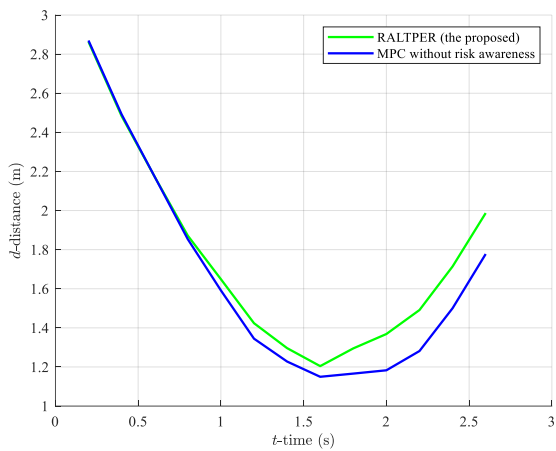

**Fig. 9.** Distance Between Vehicle Center and Pedestrian During Approach and Departure (3.8s to 6.2s)

## V. Conclusion

A risk-aware local trajectory planner for autonomous vehicles operating in complex environments with Gaussian uncertainty obstacles has been developed. The RALTPER algorithm focuses on collision avoidance constraints for both the ego vehicle region and the Gaussian-obstacle risk region. Subsequent simulations in static complex environments demonstrated the feasibility of the collision avoidance constraints. We also simulated scenarios involving a pedestrian on a narrow single-lane road and validated the algorithm's performance in dynamic environments. The experiments confirmed that the proposed algorithm effectively manages environmental uncertainty, and the novel obstacle representation extends the algorithm's applicability, offering better generalization. In future work, we will consider a broader range of uncertainty representations and conduct real-world experiments to further validate the algorithm.

## References

[1] LAVALLE S M, Rapidly-exploring random trees: a new tool for path planning[R]. Iowa State University Research Report, 19989811:1-4.
[2] KARAMAN S, FRAZZOLI E. Sampling-based algorithms for optimal motion planning[I]. The International Journal of Robotics Research, 2011, 30(7): 846-894.
[3] KUWATA Y, TEO J, FIORE G, et al. Real-time motion planning with applications to autonomous urban driving[J]. IEEE Transactions on Control Systems Technology, 2009, 17(5):1105-1118.
[4] Lei Y, Wang Y, Wu S, et al. A fuzzy logic-based adaptive dynamic window approach for path planning of automated driving mining truck[C]//2021 IEEE International Conference on Mechatronics (ICM). IEEE, 2021: 1-6.
[5] Yang H, Xu X, Hong J. Automatic Parking Path Planning of Tracked Vehicle Based on Improved A* and DWA Algorithms[J]. IEEE Transactions on Transportation Electrification, 2022, 9(1): 283-292.
[6] KIM D, CHUNG W, PARK S. Practical motion planning for car-parking control in narrow environment[J]. IET control theory &applications, 2010, 4(1): 129-139.
[7] Song G, Amato N M. Randomized motion planning for car-like robots with C-PRM[C]//Proceedings 2001 IEEE/RSJ International Conference on Intelligent Robots and Systems. Expanding the Societal Role of Robotics in the the Next Millennium (Cat. No. 01CH37180). IEEE, 2001, 1: 37-42.
[8] Dolgov D, Thrun S, Montemerlo M, et al. Path planning for autonomous vehicles in unknown semi-structured environments[J]. The international journal of robotics research, 2010, 29(5): 485-501.
[9] Klaudt S, Zlocki A, Eckstein L. A-priori map information and path planning for automated valet-parking[C]//2017 IEEE Intelligent Vehicles Symposium (IV). IEEE, 2017: 1770-1775.
[10] Sedighi S, Nguyen D V, Kapsalas P, et al. Implementing voronoi-based guided hybrid a* in global path planning for autonomous vehicles[C]//2019 IEEE Intelligent Transportation Systems Conference (ITSC). IEEE, 2019: 3845-3852.
[11] Dai S, Schaffert S, Jasour A, et al. Chance constrained motion planning for high-dimensional robots[C]//2019 International Conference on Robotics and Automation (ICRA). IEEE, 2019: 8805-8811.
[12] Dawson C, Jasour A, Hofmann A, et al. Provably safe trajectory optimization in the presence of uncertain convex obstacles[C]//2020 IEEE/RSJ International Conference on Intelligent Robots and Systems (IROS). IEEE, 2020: 6237-6244.
[13] Axelrod B, Kaelbling L P, Lozano-Pérez T. Provably safe robot navigation with obstacle uncertainty[J]. The International Journal of Robotics Research, 2018, 37(13-14): 1760-1774.
[14] Schwarting W, Alonso-Mora J, Pauli L, et al. Parallel autonomy in automated vehicles: Safe motion generation with minimal intervention[C]//2017 IEEE International Conference on Robotics and Automation (ICRA). IEEE, 2017: 1928-1935.
[15] Janson L, Schmerling E, Pavone M. Monte Carlo motion planning for robot trajectory optimization under uncertainty[M]//Robotics Research: Volume 2. Cham: Springer International Publishing, 2017: 343-361.
[16] Patil S, Van Den Berg J, Alterovitz R. Estimating probability of collision for safe motion planning under Gaussian motion and sensing uncertainty[C]//2012 IEEE International Conference on Robotics and Automation. IEEE, 2012: 3238-3244.
[17] Jasour A M, Hofmann A, Williams B C. Moment-sum-of-squares approach for fast risk estimation in uncertain environments[C]//2018 IEEE conference on decision and control (CDC). IEEE, 2018: 2445-2451.
[18] Jasour A M, Williams B C. Risk Contours Map for Risk Bounded Motion Planning under Perception Uncertainties[C]//Robotics: Science and Systems. 2019: 22-26.
[19] Han W, Jasour A, Williams B. Real-Time Tube-Based Non-Gaussian Risk Bounded Motion Planning for Stochastic Nonlinear Systems in Uncertain Environments via Motion Primitives[C]//2023 IEEE/RSJ International Conference on Intelligent Robots and Systems (IROS). IEEE, 2023: 2885-2892.
[20] Wang G B. Risk-Aware Fast Trajectory Planner for Uncertain Environments Based on Probabilistic Surrogate Reliability and Risk Contours[J]. IEEE Robotics and Automation Letters, 2022, 7(4): 12435-12442.

[21] Rasekhipour Y, Khajepour A, Chen S K, et al. A potential field-based model predictive path-planning controller for autonomous road vehicles[J]. IEEE Transactions on Intelligent Transportation Systems, 2016, 18(5): 1255-1267.
[22] Jasour A, Lagoa C. Convex chance constrained model predictive control[C]//2016 IEEE 55th Conference on Decision and Control (CDC). IEEE, 2016: 6204-6209.
[23] Zhang X, Liniger A, Borrelli F. Optimization-based collision avoidance[J]. IEEE Transactions on Control Systems Technology, 2020, 29(3): 972-983.
[24] Zhang X, Liniger A, Sakai A, et al. Autonomous parking using optimization-based collision avoidance[C]//2018 IEEE Conference on Decision and Control (CDC). IEEE, 2018: 4327-4332.
[25] He R, Zhou J, Jiang S, et al. TDR-OBCA: A reliable planner for autonomous driving in free-space environment[C]//2021 American Control Conference (ACC). IEEE, 2021: 2927-2934.
[26] Chi C, Xu X, Zhou S, et al. Adaptive Control of Path Tracking Based on Improved Pure Pursuit Algorithm[C]//International Conference on Autonomous Unmanned Systems. Singapore: Springer Nature Singapore, 2023: 56-65.
[27] Wächter A, Biegler L T. On the implementation of an interior-point filter line-search algorithm for large-scale nonlinear programming[J]. Mathematical programming, 2006, 106: 25-57.
[28] Andersson J A E, Gillis J, Horn G, et al. CasADi: a software framework for nonlinear optimization and optimal control[J]. Mathematical Programming Computation, 2019, 11: 1-36.